%% file: main.tex
\documentclass[sn-basic]{sn-jnl}

\usepackage{graphicx}%
\usepackage{multirow}%
\usepackage{amsmath,amssymb,amsfonts}%
\usepackage{amsthm}%
\usepackage{mathrsfs}%
\usepackage[title]{appendix}%
\usepackage{xcolor}%
\usepackage{textcomp}%
\usepackage{manyfoot}%
\usepackage{booktabs}%
\usepackage{algorithm}%
\usepackage{algorithmicx}%
\usepackage{algpseudocode}%
\usepackage{listings}%

\usepackage{makecell}
\usepackage{pifont}
\usepackage{adjustbox}

\newcommand{\cmark}{\ding{51}}
\newcommand{\xmark}{\ding{55}}

\theoremstyle{thmstyleone}%
\theoremstyle{thmstyletwo}%

\theoremstyle{thmstylethree}%

\begin{document}

\title[Article Title]{A Survey on Self-Supervised Learning for Non-Sequential Tabular Data}


\author*[1]{\fnm{Wei-Yao} \sur{Wang}$^{\dagger}$}\email{sf1638.cs05@nctu.edu.tw}

\author[2]{\fnm{Wei-Wei} \sur{Du}$^{\ddagger}$}\email{weiwei.du@sony.com}

\author[3]{\fnm{Derek} \sur{Xu}}\email{derekqxu@cs.ucla.edu}

\author[3]{\fnm{Wei} \sur{Wang}}\email{weiwang@cs.ucla.edu}

\author[1]{\fnm{Wen-Chih} \sur{Peng}}\email{wcpeng@cs.nycu.edu.tw}

\affil[1]{\orgname{National Yang Ming Chiao Tung University}, \city{Hsinchu}, \country{Taiwan}}

\affil[2]{\orgname{Sony Group Corporation}, \city{Tokyo}, \country{Japan}}

\affil[3]{\orgname{University of California, Los Angeles}, \city{Los Angeles}, \country{USA}}

\def\thefootnote{$\dagger$}\footnotetext{This work was done during a visiting researcher at UCLA.}\def\thefootnote{\arabic{footnote}}
\def\thefootnote{$\ddagger$}\footnotetext{This work is an independent work separated from the Sony Group Corporation.}\def\thefootnote{\arabic{footnote}}

\input{Section/0-Abstract}




\maketitle

\input{Section/1-Intro}
\input{Section/1.5Problem}
\input{Section/2.1-Pretext-Task}
\input{Section/2.2-Contrastive}
\input{Section/2.3-Hybrid}
\input{Section/3-Application}
\input{Section/4-Datasets}

\input{Section/5-Future}
\input{Section/6-Conclusion}

\bibliography{main}

\end{document}

%% file: Section/0-Abstract.tex
\abstract{
Self-supervised learning (SSL) has been incorporated into many state-of-the-art models in various domains, where SSL defines pretext tasks based on unlabeled datasets to learn contextualized and robust representations.
Recently, SSL has become a new trend in exploring the representation learning capability in the realm of tabular data, which is more challenging due to not having explicit relations for learning descriptive representations.
This survey aims to systematically review and summarize the recent progress and challenges of SSL for non-sequential tabular data (SSL4NS-TD).
We first present a formal definition of NS-TD and clarify its correlation to related studies.
Then, these approaches are categorized into three groups -- \textit{predictive learning}, \textit{contrastive learning}, and \textit{hybrid learning}, with their motivations and strengths of representative methods in each direction.
Moreover, application issues of SSL4NS-TD are presented, including \textit{automatic data engineering}, \textit{cross-table transferability}, and \textit{domain knowledge integration}.
In addition, we elaborate on existing benchmarks and datasets for NS-TD applications to analyze the performance of existing tabular models.
Finally, we discuss the challenges of SSL4NS-TD and provide potential directions for future research.
We expect our work to be useful in terms of encouraging more research on lowering the barrier to entry SSL for the tabular domain, and of improving the foundations for implicit tabular data.
}

%% file: Section/1-Intro.tex
\section{Introduction}
Supervised learning has shown outstanding performance on various machine learning tasks; however, its main hurdles lie in heavily depending on expensive human annotations and generalization bottlenecks.
With the scaling of the accessibility to unlabeled data, self-supervised learning (SSL) has demonstrated its generic and robust ability of learning contextualized information from correlations within the data.
A line of studies has shown that SSL is able to push new boundaries in various domains, including text \citep{DBLP:journals/corr/abs-2203-15556,DBLP:journals/corr/abs-2309-05463}, vision \citep{DBLP:conf/cvpr/Li0HFH23,DBLP:conf/cvpr/WooDHC0KX23}, and speech \citep{DBLP:conf/nips/BaevskiHCA21,DBLP:conf/icassp/Wu23}.
In addition, SSL illustrates strong generalizability, enabling models to adapt to tasks with limited labeled records and even unseen tasks \citep{DBLP:journals/corr/abs-2304-12210}.
The key advantage of SSL is that it reduces efforts in annotating a large amount of data, while further providing generalization ability.

\begin{figure}
  \centering
    \includegraphics[width=\linewidth]{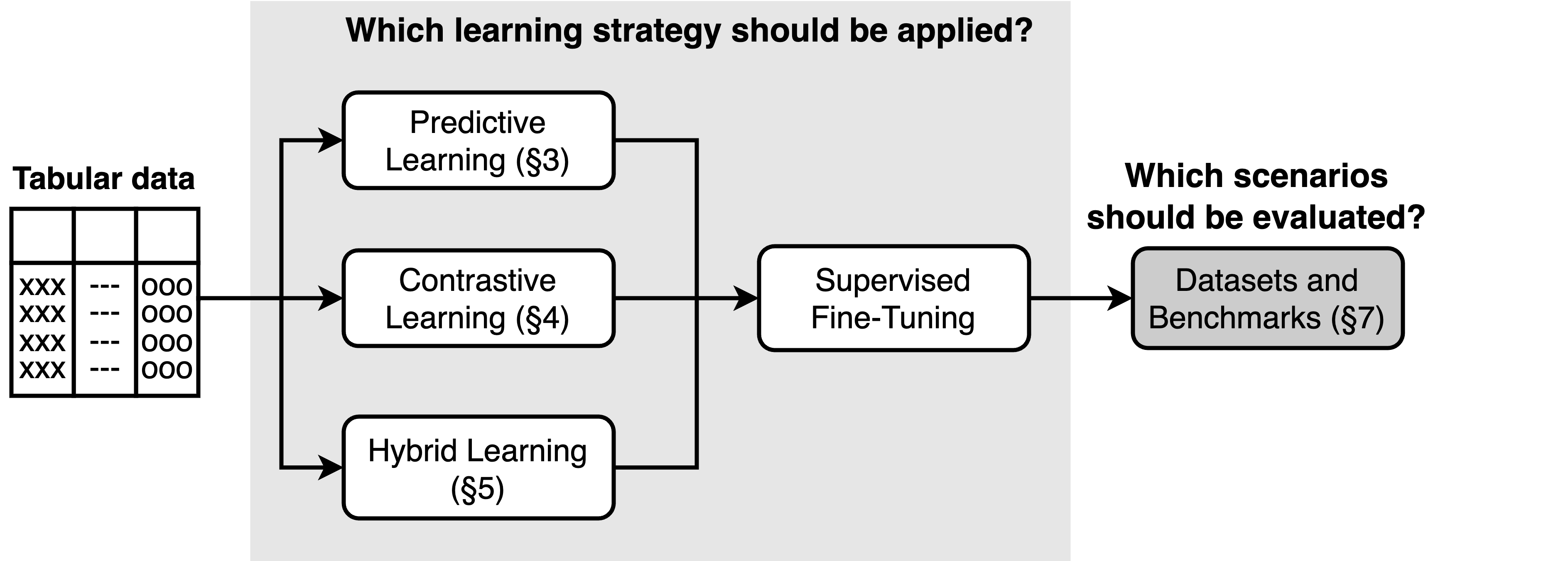}
  \caption{Overall pipeline of SSL4NS-TD. Given tabular data, the SSL4NS-TD approaches adopt predictive learning ($\S$3), contrastive learning ($\S$4), or hybrid learning ($\S$5) as the self-supervised objective before supervised fine-tuning on the downstream applications. Then, the trained model is evaluated based on the demand-related benchmarks ($\S$7), which are framed as classification or regression problems.}
  \label{fig:overview}
\end{figure}

Previous SSL methods highly relied on the unique structure of the domain datasets, such as spatial relationships in images, semantic relationships in text, and vocal relationships in speech.
In contrast, tabular data have no explicit relation between each feature, and may be completely different across various tabular datasets.
Figure \ref{fig:overview} describes the overall pipeline of learning representations from tabular data.
To opt for the learning strategy, in contrast to only using supervised learning requiring task-specific labels during the training stage, SSL is further leveraged to learn task-agnostic representations from pretext tasks, creating labels explicitly (i.e., predictive learning) and/or implicitly (i.e., contrastive learning) as categorized by \citep{DBLP:journals/tkde/LiuZHMWZT23}.
The model is expected to learn universal representations from unlabeled tabular datasets and accordingly adapt effectively to different downstream tasks including both classification and regression problems.
Generally, existing techniques with SSL for tabular data can be grouped into sequential and non-sequential tabular types for representation learning.

This survey focuses on SSL for non-sequential tabular data (SSL4NS-TD) for two reasons.
First, SSL for sequential data commonly used in fields such as recommendation systems has been widely adopted by recurrent- and Transformer-based techniques based on the temporal-ordered composition, and has already been discussed in recent survey papers \citep{badaro-etal-2023-transformers,DBLP:conf/ijcai/0001LGQZ0T23}.
On the other hand, non-sequential tabular data (NS-TD) make it more challenging to define explicit structures with pretext tasks, which has led to tremendous efforts in recent years, but has not yet been discussed in the literature.
This inspired us to provide a systematic review of recent SSL4NS-TD approaches to discuss their motivations and self-supervised objectives.
Second, whether deep learning models are superior to machine learning models in the NS-TD problems remains an active discussion topic \citep{DBLP:conf/nips/GrinsztajnOV22,DBLP:journals/corr/abs-2305-02997}.
However, SSL showcases its robustness in not only full-labeled data but also only a few labeled records, where machine learning and deep learning models fall short in such low-resource scenarios.
It is thus urgent to summarize the efforts of learning contexts with self-supervision for NS-TD to apprise the community of the current progress and to open more discussion.



The main contributions of our survey are as follows:
\begin{itemize}
    \item We present the first comprehensive survey of recent advancements in SSL4NS-TD, consisting of problem definitions, taxonomy, application issues, NS-TD datasets, and evaluation protocols. 
    Our systematic categorization of these methods into three subcategories based on their defining characteristics offers an in-depth and structured overview of the subject.
    Moreover, this survey presents an up-to-date paper survey with a paper list that will be continuously updated\footnote{A detailed list can be found at https://github.com/wwweiwei/awesome-self-supervised-learning-for-tabular-data.}.
    \item We empirically evaluate representative SSL4NS-TD approaches of each learning category on the most recent and large-scale benchmark, TabZilla \citep{DBLP:journals/corr/abs-2305-02997}, validating the most state-of-the-art SSL4NS-TD methods.
    In addition, we delve into these comparisons to provide insights into stimulating future advancements and the gap across various learning categories.
    \item We highlight the addressed challenges and future directions in SSL for the existing NS-TD methods. 
    We intend to shed light on under-researched aspects to spur further investigations that will explore the various possible paths and their interrelations within NS-TD applications, e.g., finance and healthcare.
\end{itemize}
The remainder of this review is structured as follows.
Section \ref{sec:overview} first introduces the problem of NS-TD and proposes a novel taxonomy of SSL4NS-TD consisting of three learning strategies as summarized in Table \ref{tab:taxonomy}.
The detailed discussion of the achievements, the downstream tasks, and code links (if available) of the three directions are summarized in Sections \ref{sec:pretext}, \ref{sec:contrastive}, and \ref{sec:hybrid} respectively.
Section \ref{sec:application} briefly introduces application issues (i.e., automatic data engineering, cross-table transferability, and domain knowledge integration) in practice and recent related work.
Subsequently, Section \ref{sec:datasets} describes in detail the existing benchmark datasets including both classification and regression tasks, and the benchmark results on several SSL4NS-TD approaches are reported and analyzed.
Finally, Section \ref{sec:future} sheds some light on the future research directions of SSL4NS-TD, and Section \ref{sec:conclusion} concludes this survey.

%% file: Section/1.5Problem.tex
\section{Overview}
\label{sec:overview}
\subsection{Problem Definition of SSL4NS-TD}
Tabular data consist of horizontal rows as samples and vertical columns as features of the corresponding samples, where samples and features are structured in a tabular form.
The features can be depicted with numbers, indicators, categories, text, etc.
The applications of the tabular domain can be generally formulated into two categories: \textit{classification} and \textit{regression} tasks.

In the definition of the NS-TD tasks, given a tabular dataset $D=\{x_i, y_i\}_{i=1}^{|D|}$ with $|D|$ samples, where $x_i \in X \in \mathbb{R}^{N}$ denotes a sample consisting of $N$ features, the corresponding label $y_i \in Y$ is a scalar for regression tasks or is a class for classification tasks.
The goal is to learn a predictive model $f: X \rightarrow Y$ trained by supervised loss function (e.g., cross-entropy or MSE).
Distinct from sequential tabular data, non-sequential tabular data do not have explicit chronological order (e.g., timestamp) between samples.
When applying SSL for NS-TD applications, the first intention is to construct an encoder function $e: X \rightarrow Z$, where $Z$ represents contextualized representations learned from self-supervised objectives.
It is worth noting that the labels from self-supervised objectives are obtained from the data itself instead of requiring manual annotations \citep{DBLP:journals/tkde/LiuZHMWZT23}.
The encoder function can then be used with the model in the downstream task to utilize better representations to predict the label, which is often trained either in a two-stage (i.e., pre-train then fine-tune) manner or jointly with self-supervised and downstream objectives.

\subsection{Taxonomy}
Although relatively less concentrated work has been done in SSL4NS-TD compared with other domains, various SSL4NS-TD methods have been proposed to advance the exploration of implicit tabular structures without labels, which motivated us to outline the advancements of these works.
To better understand the developing venation of SSL4NS-TD, we have identified representative and profound research works in the most influential venues as well as high-impact but preprint or workshop papers, analyzed their research motivations, and summarized their key technical contributions.
Since there is no survey literature summarizing the efforts of SSL4NS-TD, this survey establishes a novel taxonomy of SSL4NS-TD that categorizes the existing research works into three major SSL groups, as presented in Table \ref{tab:taxonomy}.
We briefly explain the core ideas of the three SSL research trends in NS-TD as follows.

\begin{table*}
    \centering
    \begin{adjustbox}{angle=90}
    \input{Tables/taxonomy}
    \end{adjustbox}
    \caption{A taxonomy for representative SSL4NS-TD algorithms with open-source codes. ``/" indicates not applicable or only the preprint version. ``Downstream" indicates the type of downstream tasks, specifically classification, regression, and both. ``P-F Dataset" indicates if an algorithm pre-trains and fine-tunes with the same downstream dataset.}
    \label{tab:taxonomy}
\end{table*}

\begin{itemize}
    \item \textbf{Predictive Learning of SSL4NS-TD} is the most widely-used category for SSL4NS-TD to benefit the downstream performance. 
    Due to the heterogeneous characteristics of features, designing prediction tasks before the final goal (i.e., downstream task) enables the model to learn background knowledge from raw data.
    The difficulty resides in devising predictive pretext tasks that are effective, considering the relations between upstream and downstream datasets and tasks.
    Although there is no consensus of designing predictive pretext tasks, several paradigms have been proposed, including learning from masked features \citep{DBLP:journals/corr/abs-2012-06678,DBLP:conf/nips/YoonZJS20,DBLP:conf/aaai/ArikP21,DBLP:conf/iclr/LeeIS22,wu2024switchtab}, perturbation in latent space \citep{DBLP:conf/iclr/NamTLLS23,sui2023selfsupervised}, and inherent in pre-trained language models \citep{DBLP:conf/nips/DinhZZLGRSP022,DBLP:conf/iclr/BorisovSLPK23,DBLP:conf/emnlp/ZhangWYJL23,anonymous2024making}. 
    \item \textbf{Contrastive Learning of SSL4NS-TD} aims to learn the similarities and discrepancies of instances in the tabular domain.
    The main advantage is that contrastive learning offers a task-agnostic learning strategy that can be applied in a wide range of downstream applications and transferability with only a few labeled samples; however, the challenge lies in the design of which instances should be closer and be pulled over.
    Researchers thus address it by adopting different views of tabular data, such as instance-wise \citep{DBLP:conf/iclr/BahriJTM22}, model-wise \citep{hajiramezanali2022stab}, column-wise \citep{DBLP:conf/nips/Wang022}, and latent space-wise \citep{anonymous2024ptarl}.
    \item \textbf{Hybrid Learning of SSL4NS-TD} extends to combine predictive learning and contrastive learning as the SSL objective, which is moving towards a more unified SSL4NS-TD.
    The principal benefit of hybrid learning is that it integrates the advantages of both learning strategies.
    As there are various successful paradigms of predictive learning, researchers have attempted to explore the effectiveness of SSL4NS-TD by the combination of perturbation and contrastive learning \citep{DBLP:conf/nips/UcarHE21,somepalli2022saint,chen2023recontab,DBLP:conf/iclr/YangWLWL24,DBLP:conf/icml/ZhuSE0KS23} as well as masking and contrastive learning \citep{DBLP:conf/cikm/DuWP23,DBLP:conf/iclr/LevinCSBBGWG23,DBLP:journals/corr/abs-2307-04308}.
\end{itemize}

%% file: Tables/taxonomy.tex
\begin{tabular}{c|llllll}
    \toprule
    \textbf{Category} & \textbf{Algorithm} & \textbf{Encoder} & \textbf{P-F Dataset} & \textbf{Downstream} & \textbf{Venue} & \textbf{Code Link (Github)} \\
    \midrule
    \multirow{12}{*}{\makecell[c]{Predictive\\Learning}}
        & VIME$^{[1]}$ & MLP & \cmark & Both & NeurIPS-20 & \tiny\href{https://github.com/jsyoon0823/VIME}{jsyoon0823/VIME} \\
        & TabTransformer$^{[2]}$ & Transformer & \cmark & Classification & / & \tiny\href{https://github.com/lucidrains/tab-transformer-pytorch}{lucidrains/tab-transformer-pytorch} \\
        & TabNet$^{[3]}$ & Transformer & \cmark & Both & AAAI-21 & \tiny\href{https://github.com/dreamquark-ai/tabnet}{dreamquark-ai/tabnet} \\
        & SEFS$^{[4]}$ & MLP & \cmark & Classification & ICLR-22 & \tiny\href{https://github.com/chl8856/SEFS}{chl8856/SEFS} \\
        & LIFT$^{[5]}$ & Transformer & \xmark & Both & NeurIPS-22 & \tiny\href{https://github.com/UW-Madison-Lee-Lab/LanguageInterfacedFineTuning}{UW-Madison-Lee-Lab/LanguageInterfacedFineTuning} \\
        & TapTap$^{[6]}$ & Transformer & \xmark & Both & EMNLP-23 & \tiny\href{https://github.com/ZhangTP1996/TapTap}{ZhangTP1996/TapTap} \\
        & TabPFN$^{[7]}$ & Transformer & \cmark & Classification & ICLR-23 & \tiny\href{https://github.com/automl/TabPFN}{automl/TabPFN} \\
        & GReaT$^{[8]}$ & Transformer & \xmark & Both & ICLR-23 & \tiny\href{https://github.com/kathrinse/be\_great}{kathrinse/be\_great} \\
        & STUNT$^{[9]}$ & MLP & \cmark & Classification & ICLR-23 & \tiny\href{https://github.com/jaehyun513/STUNT}{jaehyun513/STUNT} \\
        & SwitchTab$^{[10]}$ & Transformer & \cmark & Classification & AAAI-24 & / \\
        & LFR$^{[11]}$ & MLP & \cmark & Classification & ICLR-24 & \tiny\href{https://github.com/layer6ai-labs/lfr}{layer6ai-labs/lfr} \\
        & TP-BERTa$^{[12]}$ & Transformer & \xmark & Both & ICLR-24 & / \\
    \midrule
    \multirow{4}{*}{\makecell[c]{Contrastive\\Learning}} 
        & SCARF$^{[13]}$ & MLP & \cmark & Classification & ICLR-22 & \tiny\href{https://github.com/clabrugere/pytorch-scarf}{clabrugere/pytorch-scarf} \\
        & STab$^{[14]}$ & MLP & \cmark & Classification & NeurIPS-22 TRL & / \\
        & TransTab$^{[15]}$ & Transformer & \xmark & Classification & NeurIPS-22 & \tiny\href{https://github.com/RyanWangZf/transtab}{RyanWangZf/transtab} \\
        & PTaRL$^{[16]}$ & MLP & \cmark & Both & ICLR-24 & / \\
    \midrule
    \multirow{8}{*}{\makecell[c]{Hybrid\\Learning}} 
        & SubTab$^{[17]}$ & MLP & \cmark & Classification & NeurIPS-21 & \tiny\href{https://github.com/AstraZeneca/SubTab}{AstraZeneca/SubTab} \\
        & SAINT$^{[18]}$ & Transformer & \cmark & Both & NeurIPS-22 TRL & \tiny\href{https://github.com/somepago/saint}{somepago/saint} \\
        & ReConTab$^{[19]}$ & Transformer & \cmark & Classification & NeurIPS-22 TRL & / \\
        & /$^{[20]}$ & Both & \xmark & Classification & ICLR-23 & \tiny\href{https://github.com/LevinRoman/tabular-transfer-learning}{LevinRoman/tabular-transfer-learning} \\
        & DoRA$^{[21]}$ & MLP & \cmark & Regression & CIKM-23 & \tiny\href{https://github.com/wwweiwei/DoRA}{wwweiwei/DoRA} \\
        & CT-BERT$^{[22]}$ & Transformer & \xmark & Classification & / & / \\
        & XTab$^{[23]}$ & Transformer & \xmark & Both & ICML-23 & \tiny\href{https://github.com/BingzhaoZhu/XTab}{BingzhaoZhu/XTab} \\
        & UniTabE$^{[24]}$ & Transformer & \xmark & Both & ICLR-24 & / \\
    \midrule\midrule
    \multicolumn{7}{l}{\makecell[l]{
        $^{[1]}$\cite{DBLP:conf/nips/YoonZJS20}, $^{[2]}$\cite{DBLP:journals/corr/abs-2012-06678}, $^{[3]}$\cite{DBLP:conf/aaai/ArikP21}, $^{[4]}$\cite{DBLP:conf/iclr/LeeIS22},  $^{[5]}$\cite{DBLP:conf/nips/DinhZZLGRSP022},\\ $^{[6]}$\cite{DBLP:conf/emnlp/ZhangWYJL23}, $^{[7]}$\cite{DBLP:conf/iclr/Hollmann0EH23}, $^{[8]}$\cite{DBLP:conf/iclr/BorisovSLPK23}, $^{[9]}$\cite{DBLP:conf/iclr/NamTLLS23}, $^{[10]}$\cite{wu2024switchtab},\\ $^{[11]}$\cite{sui2023selfsupervised}, $^{[12]}$\cite{anonymous2024making}, $^{[13]}$\cite{DBLP:conf/iclr/BahriJTM22}, $^{[14]}$\cite{hajiramezanali2022stab}, $^{[15]}$\cite{DBLP:conf/nips/Wang022},\\ $^{[16]}$\cite{anonymous2024ptarl}, $^{[17]}$\cite{DBLP:conf/nips/UcarHE21}, $^{[18]}$\cite{somepalli2022saint}, $^{[19]}$\cite{chen2023recontab}, $^{[20]}$\cite{DBLP:conf/iclr/LevinCSBBGWG23},\\ $^{[21]}$\cite{DBLP:conf/cikm/DuWP23}, $^{[22]}$\cite{DBLP:journals/corr/abs-2307-04308},  $^{[23]}$\cite{DBLP:conf/icml/ZhuSE0KS23}, $^{[24]}$\cite{DBLP:conf/iclr/YangWLWL24}
    }} \\
    \bottomrule
\end{tabular}

%% file: Section/2.1-Pretext-Task.tex
\section{Predictive Learning of SSL4NS-TD} \label{sec:pretext}
As the compositions of NS-TD are heterogeneous yet without explicit relations (i.e., each column serves as a unique feature), it is challenging to distinguish relations from tables, which is one of the reasons that tree-based models are superior \citep{DBLP:conf/nips/GrinsztajnOV22}.
Motivated by SSL works in the homogeneous feature type domains (e.g., text, audio, image), such as perturbation, rotation, cropping, and adding noise, as predictive pretext tasks \citep{DBLP:journals/corr/abs-2304-12210}, tabular-based SSL objectives are mainly designed on top of these approaches to design pretext tasks.
The model is expected to be effective in downstream tasks if it is able to infer the original feature from the other masked or corrupted features.
Formally, a general SSL predictive model can be defined as:
\begin{equation}
    L_{predictive} = \psi(g(e(x^*_i)), y^*_i), 
\end{equation}
\begin{equation}
    x^*_i, y^*_i = \delta(x_i),
\end{equation}
where $\psi$ refers to the loss function that optimizes transformed input $x^*_i$ with the self-supervised label $y^*_i$, which are created by the transformed function $\delta$.
$g$ stands for the projection head aiming to convert encoded embeddings by encoder $e$ into the self-supervised prediction.


\subsection{Learning from Masked Features} 
The objective for masking features of a sample enables the model to learn the sample context via partially known features, which also aligns with the analogous objective of downstream applications to predict the corresponding category/value of a sample from the given features.
This alignment offers a trained encoder the knowledge of inferring from features of a given sample in the downstream tasks.
Inspired by Masked Autoencoder (MAE) \citep{DBLP:conf/cvpr/PathakKDDE16}, utilizing random masking of the pixels and then reconstructing them to learn numerous visual concepts, TabTransformer \citep{DBLP:journals/corr/abs-2012-06678} and VIME \citep{DBLP:conf/nips/YoonZJS20} optimized pretext models by recovering an input sample from its corrupted or masked variants, building a framework to generalize to all tabular data.
Specifically, TabTransformer introduced random masking and random value replacement as the transformed function.
VIME identified the masked features with the mask vector estimator and simultaneously imputed the masked features from the correlated non-masked features with the feature vector estimator; for example, if the value of a feature is very different from its correlated features, this feature is likely masked. 
The binary mask of VIME is randomly sampled from a Bernoulli distribution.

To further improve VIME by encouraging the encoder to generate more structured and representative embeddings, TabNet \citep{DBLP:conf/aaai/ArikP21} devised an attention mechanism to iteratively choose the features to be masked, enabling interpretability of the deep learning model.
Contrary to learnable masking, SEFS \citep{DBLP:conf/iclr/LeeIS22} proposed a feature subset generator as the transformed function by enhancing the probability of masking highly correlated features.
SwitchTab \citep{wu2024switchtab} leverages the asymmetric encoder-decoder architecture on top of the self-supervised objective of VIME, and proposes a switching mechanism to decouple two samples, each of which consists of mutual information that is switchable between samples and salient information that is unique to each sample.
Therefore, the contrastive goal is to recover the input features from both mutual and salient information.
Despite the progress, the percentage of masking presents an indecisive and case-by-case issue that every downstream task requires the percentage by empirical adjustments.

\subsection{Perturbation in Latent Space}
To learn the generalizable context from heterogeneous characteristics of tabular data, STUNT \citep{DBLP:conf/iclr/NamTLLS23} meta-learns self-generated tasks from unlabeled data, which is motivated by columns that may share correlations to the downstream labels (e.g., the feature of ``occupation" can be used as a substituted label for ``income" before the supervised learning stage).
The transformed function masks some features of a table, which are then used for k-means clustering to generate pseudo-labels.
On top of the meta-learning schema, STUNT is effective in few-shot tabular scenarios.
LFR \citep{sui2023selfsupervised} explored random projectors to learn from unlabeled data in the scenario of lacking knowledge to augment the data to serve as a universal framework that can incorporate various data modalities and domains; nonetheless, they uncovered that it is likely to be inferior for the scenario that has sufficient information for augmentations.

\subsection{Inherent in Pre-Trained Language Models}
Taking another direction to solve the feature heterogeneity issue, applying language models as an encoder is able to empower transferred knowledge across different datasets by representing tabular data with semantic text.
To address the significant challenge of transferring tabular data into a natural language format, various pre-trained language models (PLMs) have been incorporated with the NS-TD problems to leverage pre-trained knowledge from natural language corpora.
Several works \citep{DBLP:conf/nips/DinhZZLGRSP022,DBLP:conf/iclr/BorisovSLPK23,DBLP:conf/emnlp/ZhangWYJL23} have directly regarded numerical features as a string, which is intuitive but mitigates the effort of preprocessing (e.g., there is no need to transform category features into one-hot encoding).
To force the language model to interpret numerical features, TP-BERTa \citep{anonymous2024making} proposed relative magnitude tokenization to transfer scalar to discrete tokens by binning numerical values with decision trees.
Subsequently, the embeddings of binned tokens are then multiplied by the original values to avoid a large number of values within a single bin.
To eliminate the feature order bias, GReaT \citep{DBLP:conf/iclr/BorisovSLPK23} connects tabular and text modalities with a textual encoding schema and randomly permutes the order of features.
For instance, the schema converts the input tabular features ``age = 26; income = 70k" as ``age is 26, income is 70k" to be consumed by LLMs.
The order perturbation is then perturbed as ``income is 70k, age is 26" as synthetic data generation results.

%% file: Section/2.2-Contrastive.tex
\section{Contrastive Learning of SSL4NS-TD}
\label{sec:contrastive}



Another common theme of the advancements is to learn robust representations via different views or corruptions of the same input, which is achieved by maximizing similarities between similar instances and pulling over instances that are dissimilar.
With the success of generating views in computer vision (CV) and masking tokens in natural language processing (NLP) \citep{DBLP:journals/corr/abs-2011-00362}, contrastive learning has been attempted in tabular applications to learn effective and generic task-agnostic representations.
Formally, the formula of contrastive learning can be generally defined as:
\begin{equation}
    L_{contrastive} = \phi(h_i, h_i^+, h_j^-), 
\end{equation}
\begin{equation}
    h_i = e(x_i); h_i^+ = e(x_i^+); h_j^- = e(x_j^-),
\end{equation}
where $\phi$ is a similarity function that compares similarities between the anchor ($h_i$), positive ($h_i^+$), and negative ($h_i^-$) instances.
A positive instance is a variant of the anchor by augmentation methods (e.g., partial features of the anchor), and a negative instance is often sampled from the other instances.
Positive pairs represent two similar instances, while negative pairs represent two dissimilar instances.
Generally, the similarity function can be opted by cosine similarity, Euclidean distance, or dot product to compute pairwise similarity.
Note that positive and negative pairs may be optional depending on the contrastive loss as depicted next.
The projection head $g$ may need to be applied after the output of $e$ if the objective requires self-supervised labels (e.g., supervised contrastive learning).

SCARF \citep{DBLP:conf/iclr/BahriJTM22} is an MLP-based framework with a two-stage learning strategy: InfoNCE \citep{DBLP:journals/corr/abs-1807-03748} contrastive pre-training and supervised fine-tuning, which reinforces the generalizability for various tabular domains.
In the pre-training stage, the given input is corrupted with a random subset of its features, which are then replaced by a random view from the marginal distribution of the corresponding features to represent positive and negative instances.
Subsequently, InfoNCE is applied as the similarity function to encourage the sample and the variant of the corresponding sample to be close, and the sample and the variants of the other samples to be far apart:
\begin{equation}
    \phi_{InfoNCE}(h_i, h_i^+, h_j^-) = log(\frac{exp(\frac{h_i \cdot h_i^+}{\tau})}{\sum_{j=1}^N exp(\frac{h_i \cdot h_j^-}{\tau})}),
\end{equation}
where $\tau$ is the temperature parameter.

In contrast to SCARF, STab \citep{hajiramezanali2022stab} aims to introduce an augmentation-free self-supervised representation learning technique that does not require the need for negative pairs.
Specifically, STab encodes the input sample with two MLP-based encoders (one with an additional projection head), which are weight-sharing but have different stochastic regularization, which can be viewed as model-wise contrastive learning, and then compares the negative cosine distance as the similarity function.
The two representations are therefore considered as a corrupt version of the other one instead of selecting from the subset of its features.
To learn contexts across tables with disparate columns that can be used for transfer learning, feature incremental learning, and zero-shot inference, TransTab \citep{DBLP:conf/nips/Wang022} contextualizes the columns and cells in tables (e.g., gender is woman instead of using a categorized number) with Transformer encoders, and pre-trains on multiple tables with vertical-partition contrastive learning that designs variants based on column-wise splitting views.
\cite{anonymous2024ptarl} proposed a prototype-based tabular representation learning framework to learn disentangled representations around global data prototypes, which provides global prototypes to confront similar samples while preserving original distinct information with the diversifying constraint in the latent space.

%% file: Section/2.3-Hybrid.tex
\section{Hybrid Learning of SSL4NS-TD}
\label{sec:hybrid}
As predictive learning and contrastive learning of SSL4NS-TD have their own unique advantages and incorporate distinct self-supervision signals, an important learning strategy is to integrate both dimensions of SSL4NS-TD into a single model to provide multifaceted self-supervised tasks.
Typically, models equipped with hybrid learning require multiple projection heads for different pretext tasks (i.e., multi-tasking), which are employed in parallel to enhance self-supervision robustness.
Formally, the loss of hybrid learning can be defined as:
\begin{equation}
    L_{hybrid} = L_{predictive} + L_{contrastive},
\end{equation}
where it not only takes predictive signals into account but also leverages similarity-based functions to learn jointly.
Various hybrid learning techniques of SSL4NS-TD have been adopted to optimize $L_{hybrid}$, including perturbation + contrastive learning and masking + contrastive learning.

\subsection{Perturbation + Contrastive Learning}
Perturbation with contrastive learning provides a natural benefit that is able to learn robust representations without specifying the explicit knowledge of tables, and captures contextualized relations between rows, columns, and even cells.
In addition to the reconstruction loss, SubTab \citep{DBLP:conf/nips/UcarHE21} divides tabular data into multiple subsets with potentially overlapping columns as different views (similar to the idea of cropping images in CV) for contrastive loss and distance loss, both of which enable the model to move the corresponding samples in subsets closer to each other.
To perturb features unequally for feature reconstructions, SubTab adds Gaussian noise to 1) random columns, 2) a random region of neighboring columns, or 3) random features in a sample with a binomial mask.
To prevent similar features from weighing too significantly in the reconstruction loss, \cite{chen2023recontab} integrated a regularization matrix with the reconstruction loss.
With the cooperation of classification labels, they adopted different views for supervised contrastive learning based on the labels to maximize the similarity with the same categories, and leveraged semi-supervised learning to jointly pre-train the Transformer model.

In addition to the achievements of employing Transformers for tabular data, researchers have started to frame NS-TD as tokens, which have been widely used in NLP and CV domains.
Several variants have been proposed to capture fine-grained representations in tabular data (e.g., cells, numerical, categories) \citep{somepalli2022saint,DBLP:conf/icml/ZhuSE0KS23,DBLP:conf/iclr/YangWLWL24}. The major advantage is that representations can be shared across various tabular datasets and can be modeled with self-supervision.
SAINT \citep{somepalli2022saint} described a sample with a sequence composed of the corresponding categorical or numerical features, with a special token [CLS] appended at first, similar to BERT \citep{DBLP:conf/naacl/DevlinCLT19}.
To model invariant fine-grained feature representations from other similar samples, SAINT embeds categorical as well as numerical features and encodes them with intersample attention across different rows to pre-train on a reconstruction loss and InfoNCE contrastive loss with augmentations from the embedding space.

In contrast to most existing works that perform pre-training and fine-tuning per downstream dataset, another important aspect is to pre-train tabular transformers across diverse collection tables that vary in the number and types of columns.
It provides the ability to serve as foundation models on a wide range of downstream tabular applications such as ChatGPT in NLP \citep{DBLP:journals/corr/abs-2302-09419}.
XTab \citep{DBLP:conf/icml/ZhuSE0KS23} is a general tabular transformer pre-training on the large diversity cross-tables, which is flexible to leverage existing encoder backbones (e.g., \cite{DBLP:conf/nips/GorishniyRKB21,somepalli2022saint}) and existing self-supervised strategies (e.g., reconstruction loss and contrastive loss).
UniTabE \citep{DBLP:conf/iclr/YangWLWL24} pre-trains on large-scale (13 billion examples) tabular datasets across diverse domains with a Transformer encoder-decoder architecture.
The decoder takes free-form and task-specific prompts as well as contextualized representations from the encoder to adaptively reason on task-specific customizations.
That is, the prompt can be modified to adapt to specific downstream applications.
The pre-training objective of UniTabE includes multi-cell-masking to reconstruct a portion of cells of a sample and contrastive learning to treat subsets of the same sample as positive pairs and subsets of different samples as negative pairs.

\subsection{Masking + Contrastive Learning}
Another contribution to the realm of hybrid learning is to combine feature masking with contrastive learning since it combines the advantages of aligning the objectives between upstream and downstream data while preserving the task-agnostic learning strategy.
Compared with perturbations which leave partial information for the targeted features, masking strategies completely remove the targeted features that do not contain any original information.
To accommodate different features between upstream and downstream tabular data, \cite{DBLP:conf/iclr/LevinCSBBGWG23} introduced a pseudo-feature approach on top of existing deep tabular models for pre-training, which is able to predict missing features in upstream data but which are present in downstream data, and leverages a contrastive pre-training strategy, similar to \citep{somepalli2022saint}.
\cite{DBLP:journals/corr/abs-2307-04308} pre-trained a Transformer encoder with 2k high-quality cross-table datasets with masked table modeling to learn underlying relations between features and supervised contrastive learning to cluster samples with the same label.
The insights from analyses uncover that pre-training provides more transferability over tree-based baselines.
Instead of feature-agnostic SSL approaches, the motivation of DoRA \citep{DBLP:conf/cikm/DuWP23} focuses on designing a pretext task based on domain knowledge in the financial domain for real estate appraisal.
They introduced an intra-sample pretext task by selecting the domain-specific feature of a sample as the self-supervised label during the pre-training stage (i.e., predict the located town of the given real estate).
Inter-sample contrastive learning is also adopted based on contrastive learning to separate dissimilar samples based on the domain-specific feature (i.e., two real estates located in the same town are closer).

%% file: Section/3-Application.tex
\section{Tackling Application Issues of SSL4NS-TD}
\label{sec:application}


\begin{table*}[t]
    \centering
    \input{Tables/dataset}
    \caption{Existing NS-TD benchmarks and a pre-trained dataset. TabPretNet consists of the unlabeled set for pre-training and the labeled set for downstream evaluations. 
    ``C" denotes classification, ``R" denotes regression, and ``U" denotes unlabeled. ``*" indicates the numbers are only provided the average numbers in their original papers.}
    \label{tab:dataset}
\end{table*}

The advancement of SSL4NS-TD is highly application-oriented since tabular data represent ubiquitous practical utility in diverse domains, including medicine, finance, and many other areas \citep{DBLP:journals/corr/abs-2305-02997}, and research-oriented as it remains a hurdle to explore relations between rows (sample), columns (features), as well as tables (tasks).
However, the main bottlenecks of existing deep learning and machine learning models for the NS-TD applications demand human engineering efforts, require a large number of annotated labels, and struggle with generalization from known to new scenarios.
As SSL4NS-TD manifests effective performance by learning pretext tasks from unlabeled tabular data, we explore several emerging and prevalent applications of SSL4NS-TD, showcasing their potential in the following.

\subsection{Automatic Data Engineering}
Although deep learning models alleviate the burden of feature engineering compared with machine learning models, the stable performance in various tasks remains a challenge due to imbalanced, missing, and noisy data \citep{DBLP:conf/nips/GrinsztajnOV22}.
\cite{DBLP:journals/corr/abs-2012-06678} demonstrated that SSL4NS-TD has the potential to maintain robust performance in terms of these scenarios across different datasets, resulting in a reduction in manual costs with minimal engineering efforts.
\cite{DBLP:conf/iclr/LeeIS22} effectively utilized gate vector estimation to self-supervise the selection process of correlated features that explicitly avoid selecting redundant features and allow informative features to be learned.
Therefore, leveraging the capabilities of SSL4NS-TD can yield significant benefits in various data engineering applications.

\subsection{Cross-Table Transferability}
Directly learning representations from a table requires a trained model for each downstream dataset, and suffers from strict features between training and testing data, causing a severe burden in scaling and transferring to different problems.
Therefore, how to learn representations across tables has been a critical demand for NS-TD systems in real-world scenarios, which benefits from reducing efforts on engineering features based on each dataset.
Recent approaches achieving transferability involve SSL pre-training from PLMs to contextualize knowledge with coherent semantics (e.g., LIFT, TP-BERTa, GReaT) and from scratch with fine-grained feature encodings (e.g., TransTab, XTab, UniTabE).
These methods have demonstrated that pre-training with SSL4NS-TD confers advantages for adaptation to incremental columns, low-resource scenarios, and missing value predictions \citep{DBLP:conf/iclr/YangWLWL24,DBLP:conf/icml/ZhuSE0KS23}.

\subsection{Domain Knowledge Integration}
Tabular applications often have the need to incorporate expert knowledge to infer the results (e.g., clinical trials in the medical and real estate market in finance).
\cite{DBLP:conf/cikm/DuWP23} discovered that using geographic-related features as pretext tasks is the key factor in appraising the price of real estate.
\cite{DBLP:conf/iclr/NamTLLS23} designed self-generated tasks with pseudo-labels that have significant correlations with the downstream labels (e.g., predicting real estate prices through location and property size is similar to a new task that predicts rental rates by location and property size).

%% file: Tables/dataset.tex
\begin{tabular}{l|l|l|l|l}
    \toprule
    \textbf{Benchmark} & \textbf{Task Type} & \textbf{\#Datasets} & \textbf{\#Samples} & \textbf{\#Features} \\
    \midrule
    OpenML-CC18$^{[1]}$ & C & 72 & 500 - 92,000 & 5 - 3,073 \\
    DLBench$^{[2]}$ & C + R & 11 & 7,000 - 1,000,000 & 10 - 2,000 \\
    TabularBench$^{[3]}$ & C + R &  45 &  3,000 - 10,000 &  5 - 613 \\
    TabZilla$^{[4]}$ & C & 36 &  300 - 1,000,000 &  7 - 4,297 \\
    TP-BERTa$^{[5]}$ & U &  202 & 10,000 - 100,000 & 1 - 31 \\ 
    & C + R &  145 & 10 - 9,800 & 3 - 32 \\ 
    OpenTabs$^{[6]}$ & U & 2,000 & 23,000* & 24* \\ 
    UniTabE$^{[7]}$ & U & 283,000 & 46,000* &  31* \\ 
    \midrule\midrule
    \multicolumn{5}{l}{\makecell[l]{
        $^{[1]}$\cite{DBLP:conf/nips/BischlCFGHLMRV21}, 
        $^{[2]}$\cite{DBLP:journals/inffus/Shwartz-ZivA22}, $^{[3]}$\cite{DBLP:conf/nips/GrinsztajnOV22},\\ $^{[4]}$\cite{DBLP:journals/corr/abs-2305-02997},  $^{[5]}$\cite{anonymous2024making}, $^{[6]}$\cite{DBLP:journals/corr/abs-2307-04308}, $^{[7]}$\cite{DBLP:conf/iclr/YangWLWL24}
    }} \\
    \bottomrule
\end{tabular}

%% file: Section/4-Datasets.tex
\section{NS-TD Datasets and Benchmarks}
\label{sec:datasets}

\subsection{Existing NS-TD Datasets}
In addition to application issues, fair dataset selection is also important for benchmarking different NS-TD algorithms. Due to the vast diversity of tabular data, state-of-the-art algorithms are significantly impacted by dataset selections and the corresponding hyperparameter tuning. 
To this end, recent NS-TD benchmarks demonstrate evaluating SSL4NS-TD against both machine learning and deep learning algorithms on datasets of varying properties, as summarized in Table \ref{tab:dataset}.

Most NS-TD benchmarks are derived from the UCI Machine Learning Repository and the OpenML platform. Both sources feature a large collection of classification and regression NS-TD datasets. In particular, the OpenML platform released a benchmark suite, OpenML-CC18~\citep{DBLP:conf/nips/BischlCFGHLMRV21}, consisting of 72 curated classification NS-TD datasets. Despite pushes for standardization, most NS-TD and SSL4NS-TD algorithms focus on a subset of datasets from these 2 sources.

The choice of datasets is important. Issues with dataset selection among NS-TD algorithms were first introduced in DLBench~\citep{DBLP:journals/inffus/Shwartz-ZivA22}, which compared tree-based algorithms to recently proposed deep learning algorithms on 11 classification and regression datasets with 7,000 to 1,000,000 samples. These works found that XGBoost~\citep{DBLP:conf/kdd/ChenG16} outperformed more recent deep learning approaches~\citep{DBLP:conf/iclr/PopovMB20,DBLP:conf/aaai/ArikP21,Kaggle:Baosenguo,DBLP:conf/iclr/0001EE21}, when evaluated on their shared datasets. Hence, experimental results can easily be skewed by dataset selection. TabularBench~\citep{DBLP:conf/nips/GrinsztajnOV22} extended this analysis to SSL4NS-TD  \citep{DBLP:conf/nips/GorishniyRKB21,somepalli2022saint} across 45 classification and regression task datasets with 3,000 to 10,000 samples. TabularBench revealed that tree-based models yield superior predictions more consistently with much less computational cost since both deep learning and SSL4NS-TD approaches suffer from oversmoothing tabular decision boundaries, noisy features, and misrepresenting non-rotation-invariant data. However, these conclusions mainly extend to medium-sized datasets.

TabZilla~\citep{DBLP:journals/corr/abs-2305-02997} conducted the most comprehensive large-scale study against 19 machine learning, deep learning~\citep{DBLP:conf/nips/GorishniyRKB21}, and SSL4NS-TD algorithms~\citep{somepalli2022saint, DBLP:conf/aaai/ArikP21, DBLP:conf/nips/YoonZJS20, DBLP:conf/iclr/Hollmann0EH23} across 176 classification tabular datasets, and provided a benchmark of the 36 hardest datasets with 300 to 1,000,000 samples. Tabzilla uncovered \textbf{none of the existing approaches perform well on all datasets}. Particularly, transformer-based SSL4NS-TD algorithms~\citep{DBLP:conf/aaai/ArikP21} perform well on larger datasets while PFN SSL4NS-TD algorithms~\citep{DBLP:conf/iclr/Hollmann0EH23} perform well on smaller datasets. XGBoost~\citep{DBLP:conf/kdd/ChenG16} performs well on datasets with many irregular features. Among SSL4NS-TD algorithms, TabPFN~\citep{DBLP:conf/iclr/Hollmann0EH23} performs particularly well across the benchmark, definitively outperforming tree-based models with less than 1,250 samples. SAINT~\citep{somepalli2022saint} comes in second across SSL4NS-TD algorithms on the TabZilla benchmark. Tabzilla paints a fairer picture of SSL4NS-TD, where specific algorithms excel under specific dataset properties.

Previous tabular learning benchmarks have demonstrated the achievements of standardizing datasets and training setups from OpenML and UCI datasets. While most existing SSL4NS-TD algorithms obtain datasets from a mix of the OpenML platform and UCI Machine Learning Repository, others target more specialized datasets. In particular, we found that several algorithms target vision datasets~\citep{DBLP:conf/nips/UcarHE21}, synthetic datasets~\citep{DBLP:conf/iclr/LeeIS22, DBLP:conf/nips/UcarHE21, DBLP:conf/iclr/BorisovSLPK23, DBLP:conf/nips/DinhZZLGRSP022}, housing datasets~\citep{DBLP:conf/cikm/DuWP23}, and medical datasets~\citep{DBLP:conf/nips/UcarHE21, DBLP:conf/nips/YoonZJS20, DBLP:conf/nips/Wang022}. While standardized benchmarks are necessary to elucidate the data regimes each SSL4NS-TD algorithm is tailored for, application-based datasets show real-world use cases of the approach.

As the benefits of SSL4NS-TD become more apparent, effort has been placed on improving the pretraining dataset. While most SSLNS-TD algorithms perform standard train/dev/test dataset splitting, recent work curates larger and higher-quality datasets. OpenTabs~\citep{DBLP:journals/corr/abs-2307-04308} released around 2,000 unlabeled datasets. TP-BERTa~\citep{anonymous2024making, DBLP:journals/corr/abs-2307-04308} released 202 labeled and 145 unlabeled datasets derived from OpenTabs. UniTabE~\citep{DBLP:conf/iclr/YangWLWL24} released 283,000 unlabeled datasets. The open-source effort to standardize pretraining datasets further accelerates research in SSL4NS-TD.

\begin{table*}
    \centering
    \input{Tables/benchmark}
    \caption{Performance across competitive approaches of each learning group. Rankings are aggregated across 36 datasets. The Condorcet winner is computed by counting the number of pair-wise wins each algorithm has across the benchmark.}
    \label{tab:benchmark}
\end{table*}

\subsection{Evaluations of SSL4NS-TD on the Common Benchmark}
Several popular SSL tabular learning algorithms from each learning category are benchmarked on the most recent TabZilla benchmark~\citep{DBLP:journals/corr/abs-2305-02997} consisting of 36 datasets. Specifically, we experimented each algorithm for 5 optional hyperparameter optimization steps with 10-fold cross-validation under the same experimental settings as TabZilla. We set the time limit across all validation splits and hyperparameter tunings to 2 hours to reflect limited computational resources.
As not all algorithms complete in 2 hours, we report the rankings of the algorithms across a shared set of datasets, following TabZilla. Because this evaluation metric drops datasets with incomplete relative rankings~\citep{DBLP:journals/corr/abs-2405-16156}, we also evaluate by Condorcet voting \citep{DBLP:conf/allerton/WangSCK12}, which ranks models based on pair-wise ranking scores, extracting the maximum information from datasets with incomplete rankings. 

As summarized in Table~\ref{tab:benchmark}, TabPFN achieves the best average performance among SSL4NS-TD algorithms, followed by SAINT, TabNet, VIME, and TransTab.
However, the worst ranks of TabPFN and SAINT are the third rank, while TabNet is able to achieve the best performance in at least one dataset.
Hence, the optimal SSL4NS-TD algorithm remains dataset-dependent, which also indicates that there is no state-of-the-art SSL4NS-TD approach that is superior across all datasets.
Nonetheless, TabPFN receives the overall best ranks and wins, suggesting that TabPFN serves as an appropriate initial choice for consideration.
The comparison between SAINT and TabPFN illustrates that integrating predictive learning and contrastive learning may need to be further investigated.

%% file: Tables/benchmark.tex
\begin{tabular}{l|l|llll}
    \toprule
    \multirow{ 2}{*}{\textbf{Algorithm}} & \textbf{Condorcet} & \multicolumn{4}{c}{\textbf{Rank}} \\    
    & \textbf{Wins} $\uparrow$ & Mean $\pm$ Std. $\downarrow$ & Median $\downarrow$ & Min $\downarrow$ & Max $\downarrow$ \\
    \midrule
    VIME & 3 & 3.800 $\pm$ 0.600 & 4.0 & 3.0 & 5.0 \\
    TabNet & 26 & 2.700 $\pm$ 1.345 & 3.0 & 1.0 & 5.0 \\
    TabPFN & 56 & 1.500 $\pm$ 0.671 & 1.0 & 1.0 & 3.0 \\
    \midrule
    TransTab & 43 & 4.800 $\pm$ 0.400 & 5.0 & 4.0 & 5.0 \\  
    \midrule
    SAINT & 43 & 2.200 $\pm$ 0.600 & 2.0 & 1.0 & 3.0 \\ 
    \bottomrule
\end{tabular}

%% file: Section/5-Future.tex
\section{Future Directions}
\label{sec:future}

Witnessing the appreciable achievements of SSL4NS-TD, we have foreseen unresolved challenges based on the literature and the demands of real-world scenarios.
In this section, we identify and discuss key research opportunities that merit future exploration.

\subsection{A Recipe for SSL4NS-TD}
Despite the existing explorations on various pretext tasks including predictive learning, contrastive learning, and hybrid learning, the SSL techniques of most existing works are mainly motivated by the success paradigms in NLP and CV.
However, it is still unclear which SSL methods are more appropriate for the specific tabular scenarios and how to probe suitable hyper-parameters (e.g., the common masking ratio in the reconstruction pretext task and the proper batch size in contrastive learning), which are especially critical from the industrial perspective.
While there is a potential research direction that lies in designing a pretext task that is correlated with the downstream applications \citep{DBLP:conf/nips/LeeLSZ21}, more research efforts are needed to address the above problems.

\subsection{Evolution of Foundation Tabular Models}
Representation learning is a fundamental yet crucial task in tabular domains, particularly in the realm of NS-TD datasets.
Recently, foundation models such as ChatGPT have shown powerful dexterity in a variety of NLP applications, but foundation tabular models are unexplored mainly due to their heterogeneous, implicit, as well as order-invariant table characteristics.
Although some recent works have undertaken the potential effectiveness of pre-training foundation tabular models from scratch (e.g., \cite{DBLP:conf/iclr/YangWLWL24}) and from PLMs (e.g., \cite{anonymous2024making}), there still exists much exploration space for a unified foundation model that is consistently superior to both deep learning and machine learning models.
As tabular data play a vital role across diverse real-world fields, exploring foundation models for serving a diverse range of tabular applications remains an ongoing area of research.

\subsection{Continual Learning with Tabular Models}
While the transferability of LLMs has been showcased in improving downstream performance, this approach remains significantly underexplored in the context of training foundation models for non-sequential tabular data.
The challenges arise due to the immature developments of foundation tabular models as well as the diverse and heterogeneous characteristics across different tabular datasets.
A potential solution is to unify the format (e.g., JSON) of all tabular datasets to be consumed by the foundation tabular models or existing LLMs.
Nonetheless, the development of continual learning strategies for tabular models has emerged as a promising avenue for future research, with the potential to reduce resource consumption and improve the adaptation of these models to current information.

\subsection{Privacy-Aware Tabular Models}
While learning from a large volume of tabular data benefits learning representations, it is challenging to centrally train a foundation tabular model due to the sensitive and private proprietary nature of tabular applications (e.g., financial and healthcare stakeholders).
One of the potential approaches is to leverage federated learning, where multiple devices or servers collaboratively train a model without sharing their local data \citep{DBLP:conf/nips/TanLML0022}.
Instead of centralizing the data, each participating stakeholder trains the model on its own local dataset and shares only the model updates (such as gradients or weights) with a central server.
The server then aggregates these updates to improve the global model.
Nevertheless, the direct usage of federated learning in tabular applications yields challenges, primarily stemming from data quality and distributions of tabular data.
Addressing these substantial issues opens up a challenge, thereby establishing a promising research direction for the application of federated learning in tabular domains. 

\subsection{Advancements in Multi-Modal Multi-Task Environments}
While previous studies have demonstrated the effectiveness of developing deep learning models and adding SSL approaches across various tabular datasets, as reviewed in Section \ref{sec:datasets}, tree-based models are still competitive or even better options because of their compelling performance but light computational cost in terms of many-shot scenarios.
In addition, regression problems are critical in the NS-TD applications, e.g., health metrics in the medical domain and house price prediction in the financial domain.
However, most prevailing literature centers on bolstering the performance of classification problems, while omitting more challenging yet critical regression problems, which do not have definitive boundaries of labels.
Supporting multi-task abilities may augment tabular models to share common knowledge for different tasks while providing memory-efficient resources to host them.

In addition, most existing NS-TD methods have focused on tabular-based information.
However, the multi-modal integration has demonstrated effectiveness in the sequential tabular domains, e.g., training with images of the items and the corresponding metadata for recommendations \citep{DBLP:conf/www/WeiHXZ23}.
Furthermore, the conversion from tabular format to text format (e.g., 0 $\rightarrow$ seen; 1 $\rightarrow$ unseen) enables the use of leveraging generic-knowledge LLMs to learn rich context.
With the advancements of incorporating larger scale tabular data and different modalities, we believe that the investigation on SSL4NS-TD for these perspectives can highlight future research on more robust yet deployable approaches.

%% file: Section/6-Conclusion.tex
\section{Conclusion}
\label{sec:conclusion}
SSL4NS-TD serves as a ubiquitous and vital connection between deep learning and applications with implicit relations.
In this paper, we survey the existing SSL4NS-TD literature and provide an extensive review of advanced SSL4NS-TD training strategies, including predictive learning, contrastive learning, as well as hybrid learning.
Three application issues are covered to showcase the promising potential of SSL4NS-TD.
To facilitate reproducible research and compare the effectiveness of deep learning and tree-based models, we initiate the first step to summarize the representative NS-TD benchmarks and commonly used datasets for the research community, with the results of multiple recent SSL4NS-TD approaches on the large-scale benchmark.
Furthermore, we highlight critical challenges and potential directions of SSL4NS-TD for future research.
To the best of our knowledge, this is the first survey of SSL4NS-TD.
We hope this survey can highlight the current research status of SSL4NS-TD from a unified view and shed light on future work on this promising paradigm.